\documentclass[acmtog]{acmart}

\renewcommand\footnotetextcopyrightpermission[1]{}
\usepackage{booktabs} % For formal tables
\usepackage{subfigure}
\usepackage{cuted}
\usepackage{pdfpages}

\usepackage[ruled]{algorithm2e} % For algorithms

\SetAlFnt{\small}
\SetAlCapFnt{\small}
\SetAlCapNameFnt{\small}
\SetAlCapHSkip{0pt}
 
\newcommand{\ez}[1]{\textcolor{blue}{#1}}

\copyrightyear{2022}
\acmYear{2022}
\setcopyright{acmcopyright}
\acmConference{Conference Name}{Conference Date and Year}{Conference Location}
\acmDOI{10.1145/8888888.7777777}
\acmISBN{978-1-4503-1234-5/22/07}

\begin{document}
% Title portion
\title{FDNeRF: Few-shot Dynamic Neural Radiance Fields for Face Reconstruction and Expression Editing}

\author{Jingbo Zhang}
\affiliation{
 \institution{City University of Hong Kong}
}
\email{jbzhang6-c@my.cityu.edu.hk}

\author{Xiaoyu Li}
\affiliation{
 \institution{Tencent AI Lab}
}
% \email{xliea@connect.ust.hk}

\author{Ziyu Wan}
\affiliation{
\institution{City University of Hong Kong}
}
% \email{ziyuwan2-c@my.cityu.edu.hk}

\author{Can Wang}
\affiliation{
 \institution{City University of Hong Kong}
}
% \email{cwang355-c@my.cityu.edu.hk}

\author{Jing Liao}
\authornote{Corresponding author}
\affiliation{
 \institution{City University of Hong Kong, and City University of Hong Kong Shenzhen Research Institute}}
\email{jingliao@cityu.edu.hk}
\renewcommand\shortauthors{Zhang, J. et al}

\begin{teaserfigure}
\includegraphics[width=0.99\textwidth]{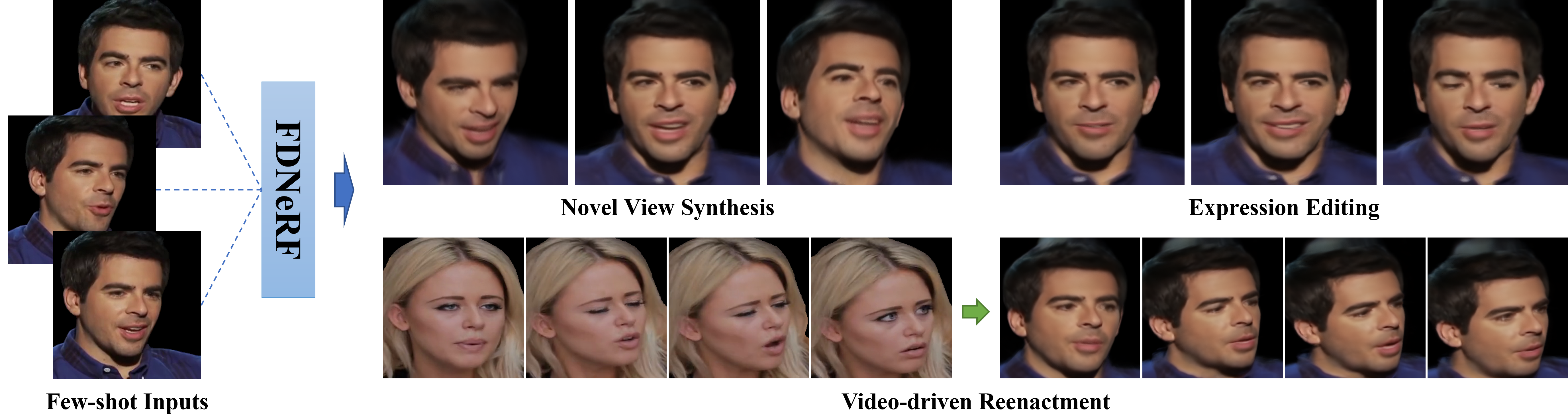}
% \vspace{-0.1in}
\caption{FDNeRF, a NeRF-based method for 3D face reconstruction with only few-shot dynamic input frames (e.g., 3 frames), enable novel view synthesis, expression editing, and video-driven reenactment tasks. Input images are from the VoxCeleb dataset \cite{nagrani2017voxceleb}.}
\label{fig:teaser}
\end{teaserfigure}

\begin{abstract}
We propose a Few-shot Dynamic Neural Radiance Field (FDNeRF), the first NeRF-based method capable of reconstruction and expression editing of 3D faces based on a small number of dynamic images. Unlike existing dynamic NeRFs that require dense images as input and can only be modeled for a single identity, our method enables face reconstruction across different persons with few-shot inputs. Compared to state-of-the-art few-shot NeRFs designed for modeling static scenes, the proposed FDNeRF accepts view-inconsistent dynamic inputs and supports arbitrary facial expression editing, i.e., producing faces with novel expressions beyond the input ones. To handle the inconsistencies between dynamic inputs, we introduce a well-designed conditional feature warping (CFW) module to perform expression conditioned warping in 2D feature space, which is also identity adaptive and 3D constrained. As a result, features of different expressions are transformed into the target ones. We then construct a radiance field based on these view-consistent features and use volumetric rendering to synthesize novel views of the modeled faces. Extensive experiments with quantitative and qualitative evaluation demonstrate that our method outperforms existing dynamic and few-shot NeRFs on both 3D face reconstruction and expression editing tasks. 
Code is available at \ez{\href{https://github.com/FDNeRF/FDNeRF}{https://github.com/FDNeRF/FDNeRF}}.

\end{abstract}


\maketitle

% ----------------------------- Introduction --------------------------
\section{Introduction}

Reconstructing and editing a human face from a small number of frames in a monocular video is a highly challenging problem in the field of computer vision and computer graphics \cite{hong2021headnerf, sun2022fenerf, sun2022ide}. Unlike reconstructing a rigid object, a faithful reconstruction of a dynamic human face is difficult because of complex geometry and appearance variations brought by rich expressions. Moreover, it is even more challenging to capture consistent multi-view frames from a monocular camera for reconstruction, which usually relies on dense synchronized cameras~\cite{gotardo2018practical, tewari2019fml, yang2020facescape, zhang2022adaptive}. 

To simplify the face reconstruction,~\citet{blanz1999morphable} proposes to represent the human face with a parametric 3D Morphable Model (3DMM), which decomposes the face attributes into low-dimensional vectors. These vectors can be used to reconstruct 3D textured face mesh using corresponding blend shapes. Based on this model, some methods~\cite{deng2019accurate, gecer2019ganfit, ploumpis2020towards} are able to reconstruct 3D facial meshes from single or few-shot images, which facilitates free-view synthesis and expression editing. However, due to the inaccurate mesh model and the limited representation ability of the low-dimensional parameters, these methods struggle to capture fine-scale details of the human face in input images, such as beards and hairs.

Recently, Neural Radiance Field (NeRF) \cite{mildenhall2020nerf}, which implicitly models the geometries and appearances of static 3D scenes as multilayer perceptrons (MLPs), has attracted widespread attention due to its impressive free-view results in photo-realistic rendering. To extend it to dynamic scenes, some dynamic NeRFs~\cite{pumarola2021d, park2021nerfies, ma2022neural} introduce a deformation field to handle the inconsistency among different frames. They can be applied to reconstruct a face in a monocular video with different expressions but require hundreds or thousands of input frames for training, which somehow limits practical usage. On the other hand, some few-shot NeRFs~\cite{gao2020portrait, raj2021pixel, hong2021headnerf} explore how to produce a generalized model that can be used to reconstruct a 3D face with only single or multi-view images. However, they require view-consistent input and cannot handle dynamic frames from a monocular video.
 
To address the challenges of modeling 3D faces with NeRFs based on few-shot dynamic frames, one possible solution is to combine the best of dynamic NeRFs and few-shot NeRFs by integrating the deformation field into an existing few-shot NeRF such as PixelNeRF \cite{yu2021pixelnerf}. However, training this 3D deformation field for human faces usually relies on a large number of images with different expressions. Moreover, the deformation fields varying across different persons, even for the same expression, impose additional challenges. Therefore, it is a severe ill-posed problem to learn a 3D deformation field conditioned on expressions and adapted to different identities, with only a small number of dynamic frames as input. To solve this problem, unlike previous dynamic NeRFs performing the 3D deformation, we propose a 2D deformation strategy in the deep feature space with 3D constraints, which is easier to be learned with few-shot frames.

In this paper, we propose a Few-shot Dynamic NeRF (FDNeRF), the first framework to reconstruct and edit 3D faces based on a small number of dynamic frames extracted from a monocular video. Our FDNeRF employs a novel Conditional Feature Warping (CFW) module with 3D constraints to handle the inconsistencies between dynamic frames by warping source expressions to the target one in the 2D feature space. Then, a reconstruction module is adopted to predict the color and density of spatial points in the radiance field based on the warped feature spaces. Finally, the volumetric rendering is used to render the results with the desired expression. 
Compared to the 3D deformation field adopted in many dynamic NeRFs~\cite{pumarola2021d, tretschk2021non, park2021nerfies, park2021hypernerf}, our CFW module implemented in the 2D feature space offers two-fold advantages. First, a 2D warping with a lower degree of freedom makes it more friendly to few-shot inputs than a 3D deformation. Second, different from the previous 3D deformation field defined on spatial positions, our 2D warping field defined on whole-image features enables it to better distinguish individuals and thus produce adaptive warping fields for different identities. Moreover, unlike conventional image warping, our feature warping of different frames is constrained by the 3D radiance field, making the 2D warping view consistent. Benefiting from the capability of the CFW module, FDNeRF can not only reconstruct 3D faces from a small number of dynamic frames but also enable editing of facial expressions and rendering of novel views. Extensive experiments demonstrate our superiority over existing methods both qualitatively and quantitatively.

% Paragraph 5: Contributions
In summary, the main contributions of this work are:
% itemize
\begin{itemize}
\item We propose FDNeRF, the first neural radiance field to reconstruct 3D faces from few-shot dynamic frames.
\item We introduce the novel CFW module to perform expression conditioned warping in 2D feature space, which is also identity adaptive and 3D constrained. 
\item Our FDNeRF supports free edits of facial expressions, and enables video-driven 3D reenactment.
\end{itemize}

\begin{figure*}
  \includegraphics[width=1\linewidth]{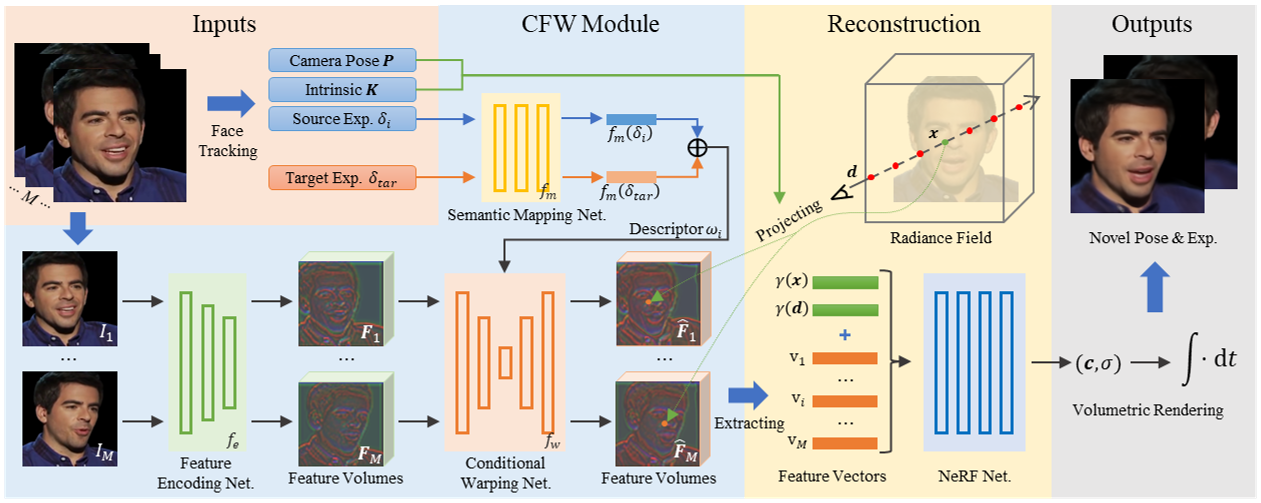}
  \vspace{-0.3in}
  \caption{Overview of our FDNeRF. Given few-shot dynamic images, face tracking is implemented to estimate relevant expression parameters $\delta_i$, camera poses $\textbf{P}$, and intrinsic matrix $\textbf{K}$ in the preprocessing stage. In the CFW module, we employ feature encoding network $f_e$ to extract a deep feature volume $F_i$ for each image $I_i$, and semantic mapping network $f_m$ to generate a motion descriptor $\omega_i$ based on source and target expression parameters. The descriptor is then used to guide the conditional warping network $f_w$ to produce warped feature volumes $\hat{F}_i$. During reconstruction, we project the query point $x$ to each image plane and extract aligned feature vectors $v_i$. These vectors, along with the position and direction of the point, are fed into NeRF network to infer color and density. Finally, volumetric rendering is performed to synthesize novel view images. Input images are from the VoxCeleb dataset \cite{nagrani2017voxceleb}.}
  \vspace{-0.1in}
  \label{fig:pipeline}
\end{figure*}

\section{Related Work}
Compared with earlier methods that represent 3D faces as parametric textured mesh models \cite{deng2019accurate}, implicit representation methods have recently received increasing attention for their impressive rendering quality on free-view synthesis \cite{or2022stylesdf, lombardi2019neural}. Notably, NeRF \cite{mildenhall2020nerf} employs MLP to learn the radiance field of a 3D scene and uses volumetric rendering to visualize the scene. Subsequently, a number of following works are proposed to extend NeRF to different scenarios, which include dynamic NeRFs aimed at alleviating the static input constraint of NeRF, and few-shot works aimed at alleviating the dense input constraint of NeRF. Since our FDNeRF focuses on modeling 3D faces from few-shot dynamic inputs, it is closely related to recent work on dynamic NeRFs and few-shot NeRFs. To clearly distinguish FDNeRF from these methods, we discuss them below.

\subsection{Dynamic NeRFs}
The vanilla NeRF assumes that the modeled scene is static and cannot reconstruct dynamic scenes from a set of frames. To solve this problem, methods like NeRFlow~\cite{du2021neural}, NSFF~\cite{li2021neural}, and DynNeRF\cite{gao2021dynamic} focus on time-varying scenes and learn 3D scene flow between two neighboring frames.
Video-NeRF \cite{xian2021space} learns a spatiotemporal irradiance field conditioned on time for dynamic scenes.
However, these methods are inappropriate for reconstructing dynamic faces from arbitrary unordered inputs since they rely on time-dependent information for reconstruction.
Hybrid-NeRF \cite{wang2021learning} introduces an additional discrete 3D-structure-aware grid of animation codes to encode dynamical properties of scenes. However, it requires multi-view videos as input and cannot be applied to monocular video-based reconstruction. AD-NeRF~\cite{guo2021ad} and NeRFace~\cite{gafni2021dynamic} form dynamic radiance fields by directly conditioning NeRF with tracked audio features or facial expressions to handle variations between different frames. Although they can achieve dynamic face modeling, they require many more frames (nearly 5k frames) than vanilla NeRF to incorporate audio or expression conditions. By contrast, D-NeRF \cite{pumarola2021d}, NR-NeRF \cite{tretschk2021non} and Nerfies \cite{park2021nerfies} propose a deformation field conditioned on spatial points and frame-related latent codes. The introduction of a 3D deformation field reduces the training difficulty of the model to a certain extent. Still, they require hundreds of frames as input to fitting a dynamic scene. Furthermore, HyperNeRF \cite{park2021hypernerf} adds an ambient slicing network to enhance the performance of Nerfies in the cases of topological changes. Although Nerfies and HyperNeRF can successfully interpolate expressions between frames, they cannot be edited to desired expressions out of the input domain.
Besides, the requirement of dense input frames for these methods still greatly limits the training speed and practical usage. 
Unlike previous dynamic NeRFs, our FDNeRF allows dynamic 3D face reconstruction with only few-shot frames.

\subsection{Few-shot NeRFs}
Simultaneously, there is another line of work that focuses on the 3D reconstruction of static scenes from a small number of input images. For example, Portrait-NeRF \cite{gao2020portrait} pretrains a canonical facial NeRF over a set of multi-view face datasets, and reconstructs a 3D face by finetuning the pretrained model on a specific facial image. HeadNeRF \cite{hong2021headnerf} and MofaNeRF \cite{zhuang2021mofanerf} propose parametric NeRF models conditioned on the 3DMM parameters extracted from input images. Although they enable expression editing by adjusting the associated 3DMM parameters, they cannot recover some facial details in the original frames due to the limited representation of low-dimensional parameters. On the other hand, some few-shot methods condition NeRF on image or feature inputs to learn a scene prior for a sparse set of inputs, like PixelNeRF \cite{yu2021pixelnerf}, PVA \cite{raj2021pixel}, and MVSNeRF \cite{chen2021mvsnerf}. Here, PixelNeRF and PVA construct radiance fields by using the implicit spatial information in the features of sparse inputs. MVSNeRF employs earlier multi-view stereo methods to produce a geometry-aware feature volume, and deduces radiance fields of target scenes based on the sampled features from this volume. Although these methods allow reconstructing photorealistic 3D scenes from a few static view-consistent images, they cannot handle the dynamic cases. By contrast, our FDNeRF combines the best of both dynamic NeRFs and few-shot NeRFs and thus enables model 3D faces from few-shot dynamic frames.

\section{Method}
\label{sec:method}
Our method enables 3D face reconstruction and expression editing based on few-shot dynamic frames (e.g., 3 frames). To this end, we propose FDNeRF, a NeRF-based dynamic face reconstruction framework, to handle inconsistencies among different frames.

\subsection{Overview}
Unlike the previous NeRF-based methods designed for dynamic scenes with dense frames, which require complex optimization for a single scene, our method tries to infer the arbitrary dynamic 3D faces using several inputs only. To accomplish this, the input frames with different facial expressions are first aligned in their 2D feature spaces via a conditional feature warping (CFW) module to eliminate the inconsistency of expressions (Sec.~\ref{sec:fwm}). 
Instead of deducing neural radiance fields based solely on positional and directional information of spatial query points like NeRF, we construct radiance fields directly from the aligned features of all input views, 
which allows us to infer the color and density of query points across identities from the radiance field (Sec.~\ref{sec:rm}).
The derived density and color along the camera rays are subsequently employed for volumetric rendering to render the final frame under novel views (Sec.~\ref{sec:vr}). The optimization procedure is lastly introduced (Sec.~\ref{sec:op}).

\subsection{3D Constrained Conditional Warping} 
\label{sec:fwm}
Given few-shot dynamic frames captured from a monocular video of a talking human head, it is hard to reconstruct the 3D face by directly using previous few-shot NeRFs designed for static scenes due to the inconsistency of facial expression among different frames. To solve this issue, one potential strategy is to optimize a deformation field to achieve 3D warping between observation and canonical spaces like existing dynamic NeRFs \cite{park2021nerfies,park2021hypernerf}. Nonetheless, the high freedom of 3D deformation requires abundant inputs of a specific person for training, which limits the construction of deformation fields across identities with few-shot inputs. 
Another naive strategy is to warp the facial expression of each frame into the same one at the image level by using existing 2D expression warping methods \cite{ren2021pirenderer, meshry2021learned}. However, without 3D constraints, the per-frame warped images lack view consistency, which would critically tamper with the following 3D reconstruction. 

To avoid the inconsistency of facial details among warped frames, we design a 2D feature warping module conditioned on expression and at the same time constrained by the 3D geometry. As shown in Fig. \ref{fig:pipeline}, the conditioned feature warping (CFW) module consists of three sub-networks: a ResNet-like feature encoding network $f_e$, a semantic mapping network $f_m$, and a conditional warping network $f_w$.
More specifically, the encoding network $f_e$ is employed to get a deep feature volume $F_i$ for each input frame $I_i$, which encodes the identity and expression information in $I_i$.
\begin{equation}
\label{eqn:encoding}
F_i = f_e \left( I_i \right),
\end{equation}
where $F_i$ is composed of feature maps extracted from the first four layers of $f_e$.

Semantic conditions to guide the warping of feature volume $F_i$ are extracted by the semantic mapping network $f_m$. Specifically, we first leverage off-the-shelf face tracking method \cite{thies2016face2face} to estimate the expression parameters $\delta$, face pose $\textbf{P}$, and intrinsic matrix $\textbf{K}$ for each input frames.
Subsequently, $f_m$ transfers the original parameters into latent codes $f_m \left( \delta_i \right)$ and $f_m \left( \delta_{tar} \right)$ for extracting more discriminative representations to achieve a fine-grained guidance in the warping network.
% Subsequently, the semantic mapping network transfers the original parameters into latent codes $f_m \left( \delta_i \right)$ and $f_m \left( \delta_{tar} \right)$ for extracting more discriminative representations to achieve a fine-grained control. 
Here, the target semantic indicates the desired expression in reconstructed face model. For each unaligned frame, we concatenate its latent code $f_m \left( \delta_i \right)$ with the target expression code $f_m \left( \delta_{tar} \right)$ to form the high-dimensional motion descriptor $\omega_i$ to guide the warping network:
\begin{equation}
\label{eqn:mapping}
\omega_i = f_m \left( \delta_i \right) \oplus f_m \left( \delta_{tar} \right).
\end{equation}

We implement the conditional warping network $f_w$ with an encoder-decoder like architecture. To more adequately guide the warping network, we inject the motion descriptor $\omega_i$ into all convolutional layers of $f_w$ by the adaptive instance normalization (AdaIN) operator. More specifically, a light-weight mapping network will transfer the motion descriptor $\omega_i$ into affine parameters $\gamma^{\omega_i}$ and $\beta^{\omega_i}$ respectively. The intermediate feature $z$ of each convolutional layer is modulated as follows:
\begin{equation}
    \operatorname{AdaIN}(z ; \omega_i)=\gamma^{\omega_i}\left(\frac{z-\mu(z)}{\sigma(z)}\right)+\beta^{\omega_i},
\end{equation}
where $\mu(\cdot)$ and $\sigma(\cdot)$ calculate the average and variance statistics regarding $z$.
Based on the feature volume $F_i$ and the descriptor $\omega_i$, the warping network $f_w$ will estimate a deformation flow field, which indicates the coordinate offsets between the input feature volume $F_i$ and the desired target feature volume $\hat{F}_i$. Ultimately we obtain the aligned feature volumes through bilinear interpolation sampling:
\begin{equation}
\label{eqn:warping}
\hat{F}_i = \textit{Sample} \left( F_i, f_w\left( F_i, \omega_i \right)\right),
\end{equation}
where $f_w\left( F_i, \omega_i \right)$ indicates the deformation flow field estimated by the warping network, $\textit{Sample} \left( a, b\right)$ represents the interpolated sampling operation on $a$ according to the flow field $b$.

It is noteworthy that although the warping fields are established in individual frames, they are actually constrained by the 3D geometry represented in the jointly trained neural radiance field, which would effectively enhance the consistency of warping across different views. We will introduce more details about the radiance fields in the following section.

\subsection{Radiance Field Reconstruction}
\label{sec:rm}
% fully-connected ResNet structure
To reconstruct the 3D face with desired expression from the target feature volumes $\hat{F}$, we adopt a framework similar to \cite{yu2021pixelnerf} as our reconstruction module to deduce the color and density of each spacial point, and then use volumetric rendering to produce the final geometry and appearance. 

Specifically, we first cast camera rays through each pixel of the target view and sample $N$ points along each ray for volumetric rendering \cite{mildenhall2020nerf}. Then, we project each sampled point $p$ on the rays to each frame coordinate using known intrinsic matrix $\textbf{K}$ and corresponding pose $\textbf{P}_i$, and extract the associated aligned feature vectors $\textbf{v}_i$ from the target feature volumes $\hat{F}_i$ via bilinear interpolation.
\begin{equation}
\label{eqn:extract}
\textbf{v}_i = \Pi \left( F_i, \textbf{K} \cdot \textbf{P}_i^{-1} \cdot \overline{\textbf{x}} \right),
\end{equation}
where $\Pi$ represents the extraction procedure, and $\textbf{K} \cdot \textbf{P}_i^{-1} \cdot \overline{\textbf{x}}$ indicates the coordinate on the $i$-th frame plane. Note that, $\overline{\textbf{x}}$ is the homogeneous coordinate of the point $p$.
% , and $\textbf{P}_i$ is the camera-to-world translation matrix of the input view $i$.

The feature vectors, as well as the position $\textbf{x}$ and direction $\textbf{d}$ of the query point $p$, are fed into the reconstruction module to estimate the color $\textbf{c}$ and density $\sigma$ values:
\begin{equation}
\label{eqn:nerf}
\left( \textbf{c}, \sigma \right)= f_\theta \left(\gamma(\textbf{d}), G(\gamma(\textbf{x}), \textbf{v}_1, \dots, \textbf{v}_M) \right),
\end{equation}
where $\gamma(\cdot)$ is the positional encoding introduced by \cite{mildenhall2020nerf} that maps the input into a higher dimensional Fourier space, and $G(\cdot)$ is the averaging function formed by a neural network to gather all available information. $M$ indicates the number of input frames, which is not fixed and can be set flexibly by the user. 
Note that, to eliminate the geometric discrepancy between views, we do not feed the direction component $\gamma(\textbf{d})$ at the beginning of the NeRF network like \cite{yu2021pixelnerf} to affect the density-related parameters. Instead, we input it into the last several layers to adjust color-related parameters only.

\begin{figure*}
  \includegraphics[width=1\linewidth]{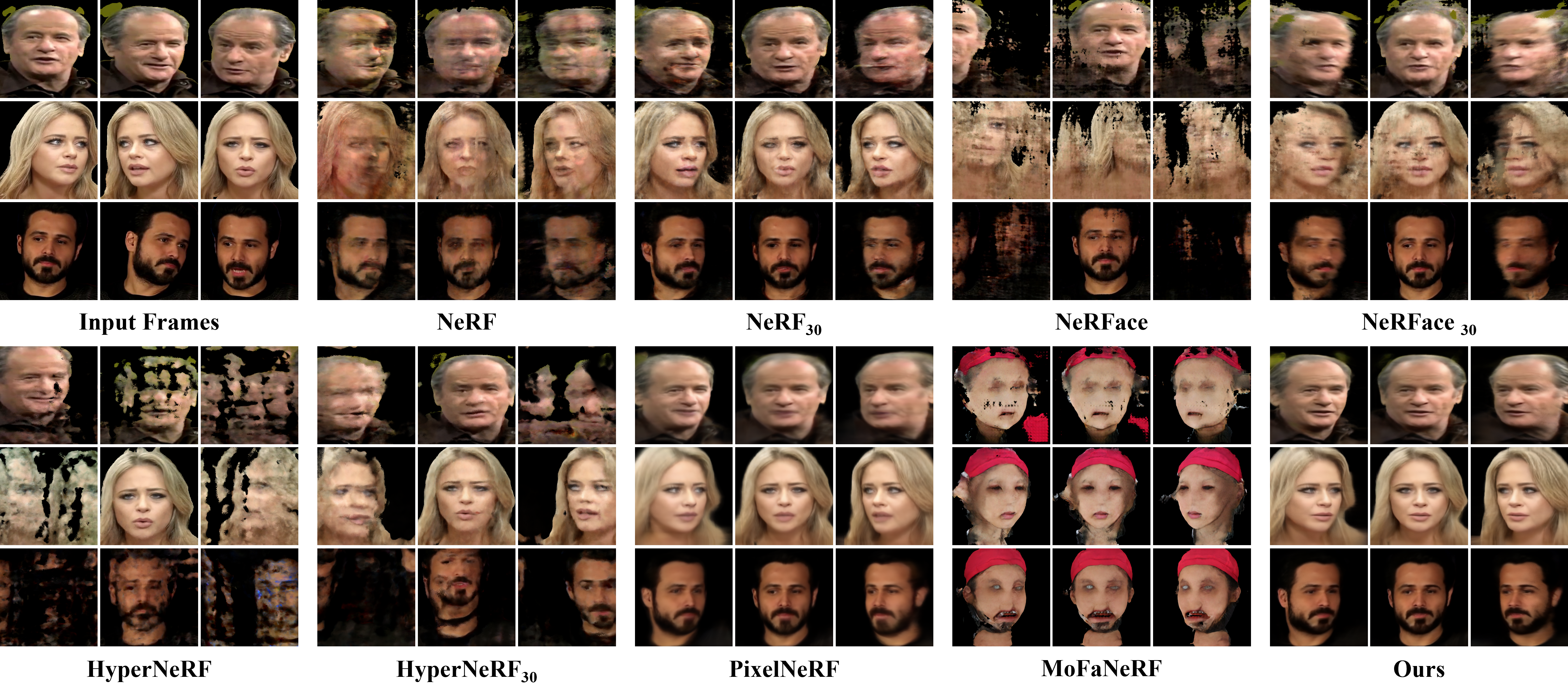}
  \vspace{-0.25in}
  \caption{Visual comparison of reconstructed 3D faces produced by baseline methods and ours. Input images are from the VoxCeleb dataset \cite{nagrani2017voxceleb}.}
  \vspace{-0.1in}
  \label{fig:rec}
\end{figure*}

\subsection{Volumetric Rendering}
\label{sec:vr}
Since our reconstruction module acts as a radiance field, we employ volumetric rendering to visualize the geometry and appearance of the implicit 3D face in a desired view. Like previous NeRF works, the expected color $\hat{\textbf{C}}$ of each pixel in the rendered image can be calculated by accumulating the estimated color $\textbf{c}$ and density $\sigma$ of all sampled points along the camera ray $\textbf{r}$:
\begin{equation}
\label{eqn:render}
\hat{\textbf{C}}(\textbf{r}) = \int_{t_n}^{t_f} T(t) \cdot \sigma \left(\textbf{r}(t)\right) \cdot \textbf{c}\left(\textbf{r}(t), \textbf{d}\right)\, dt,
\end{equation}
where $\textbf{r}(t)=\textbf{o}+t\textbf{d}$ indicates the point positions of the ray from camera center $\textbf{o}$. $t_n$ and $t_f$ are near and far bounds of the ray, respectively. $T(t)=\exp \left( -\int_{t_n}^{t}\sigma \left(\textbf{r}(s)\right)\, ds\right)$ is the accumulated transmittance along the ray.
Here, a hierarchical sampling strategy similar to \cite{mildenhall2020nerf} is adopted for efficient rendering in practice. Specifically, there are two NeRF network for coarse and fine reconstructions. The densities estimated by the coarse one are used for important sampling of query points in the fine one.

\subsection{Optimization}
\label{sec:op}
We jointly optimize the network weights of our CFW module and reconstruction module based on the photometric reconstruction loss: 
\begin{equation}
\label{eqn:loss}
\mathcal{L} = \sum_{\textbf{r}\in \mathcal{R}(\textbf{P})} \left\| \hat{\textbf{C}}(\textbf{r}) -\textbf{C}(\textbf{r}) \right\| _2^2,
\end{equation}
where $\mathcal{R}(\textbf{P})$ indicates the set of camera rays in pose $\textbf{P}$ and $\textbf{C}(\textbf{r})$ represents the pixel color in the target image. During optimization, we randomly select $M$ frames from one of the training videos as input frames and one frame from the rest frames as the target. Then, the expression parameters of input and target frames are used to guide the feature warping process in the CFW module. Note that, to adapt our model to the modeling of flexible input frames, we randomly set $M$ in the range of 1 to 12 at each optimization iteration. Besides, in order to make the optimization converge effectively, we initialize the feature encoding network in our CFW module with the weights of ImageNet pre-trained ResNet34. Apart from that, other networks in our framework are trained from scratch.

\begin{table*}[h]%
\caption{Quantitative evaluation.}
\vspace{-0.1in}
\label{tab:metric}
\begin{minipage}{\linewidth}
\begin{center}
\scalebox{0.99}{
\begin{tabular}{lcccccccc}
  \toprule
  Methods & NeRF & NeRF$_{30}$ & NeRFace & NeRFace$_{30}$ & HyperNeRF & HyperNeRF$_{30}$ & PixelNeRF &  FDNeRF \\
  \midrule
  PSNR $\uparrow$    & 17.368 & 21.963 & 13.212 & 19.454 & 10.422 & 13.521 & 24.149 & $\boldsymbol{24.847}$ \\
  SSIM $\uparrow$    & 0.537  & 0.704  & 0.281  & 0.585  & 0.252  & 0.432  & 0.792  & $\boldsymbol{0.821}$  \\
  LPIPS $\downarrow$ & 0.320  & 0.167  & 0.566  & 0.307  & 0.687  & 0.501  & 0.190  &  $\boldsymbol{0.142}$  \\
  \bottomrule
\end{tabular}
}
\end{center}
\end{minipage}
\end{table*}%

\begin{figure*}
  \includegraphics[width=1\linewidth]{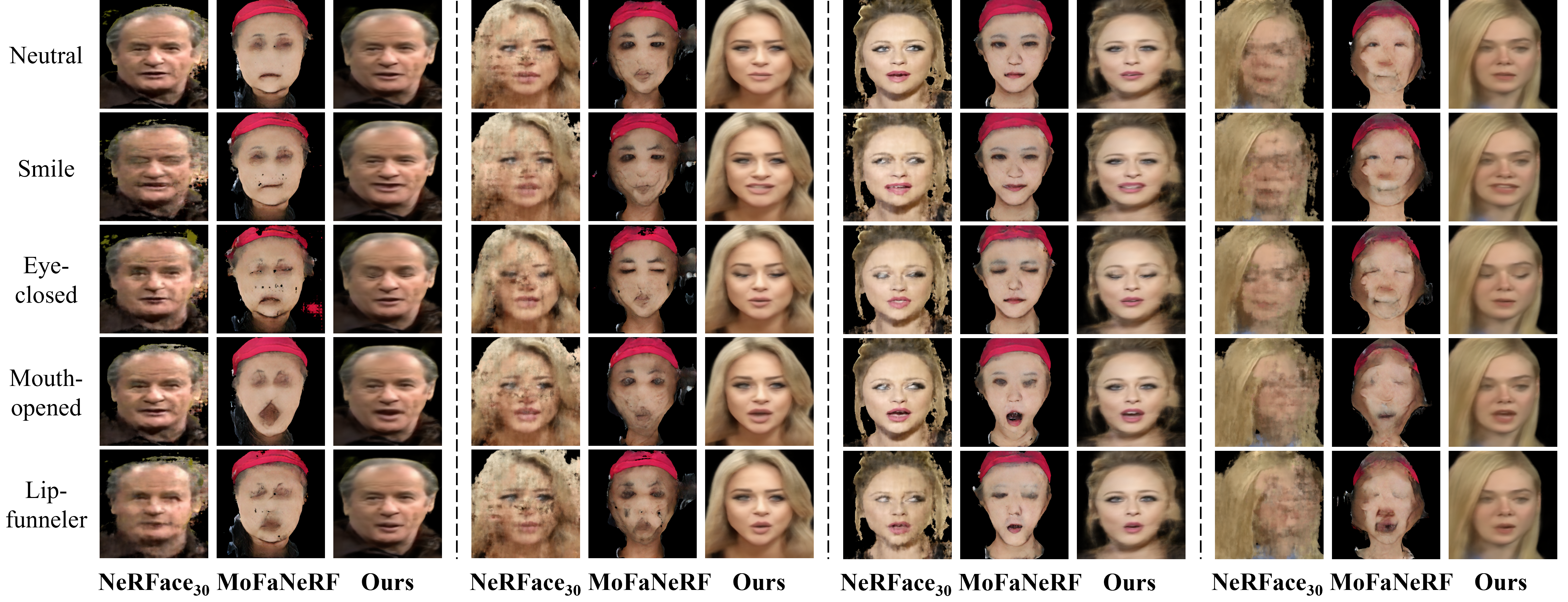}
  \vspace{-0.25in}
  \caption{Comparison on expression editing. Input images are from the VoxCeleb dataset \cite{nagrani2017voxceleb}.}
  \label{fig:exp_edit}
\end{figure*}

\section{Experiments}
\label{sec:exp}

In this section, we first give the implementation details, baselines, and metrics of our work (Sec.~\ref{sec:detail} and Sec.~\ref{sec:settting}), and then we compare our method with state-of-the-art NeRF-based methods for 3D face reconstruction and expression editing (Sec.~\ref{sec:exp_rec} and Sec.~\ref{sec:exp_edit}). Furthermore, we extend our method to implement video-driven 3D reenactment task (Sec.~\ref{sec:exp_exten}). For more experiments and ablation studies please refer to the supplementary material.

\subsection{Implementation Details}
\label{sec:detail}
We leverage Pytorch framework \cite{paszke2019pytorch} to implement FDNeRF and use Adam \cite{kingma2014adam} optimizer with default hyperparameters and a learning rate of 0.0001 to update network parameters. Our training data involves 213 talking videos which are selected from the VoxCeleb dataset \cite{nagrani2017voxceleb}.
To process these videos, we adopt the monocular face tracking method \cite{thies2016face2face} to estimate the expression semantic, face pose, and intrinsic matrix for each frame.
Here, the estimated face pose is used as the camera pose of the image in our implementation.

\subsection{Setup}
\label{sec:settting}
\noindent\textbf{Baseline methods.} ~~ For 3D face reconstruction and free-view synthesis, we compare with the vanilla NeRF \cite{mildenhall2020nerf}, two dynamic NeRFs (NeRFace \cite{gafni2021dynamic} and HyperNeRF \cite{park2021hypernerf}), and two few-shot NeRFs (MoFaNeRF \cite{zhuang2021mofanerf} and PixelNeRF \cite{yu2021pixelnerf}). Here, NeRFace and HyperNeRF are two methods for modeling dynamic scenes with dense inputs, while MoFaNeRF and PixelNeRF are designed for few-shot modeling of static scenes. We do not compare to PVA \cite{raj2021pixel} (a few-shot NeRF for face modeling) as their code is unavailable. Instead, we compare with PixelNeRF, which has a similar framework and performance to PVA.

\noindent\textbf{Evaluation metrics.} ~~For the quantitative comparisons, we employ peak-signal-to-noise ratio (PSNR), structural similarity index measure (SSIM) \cite{wang2004image}, and learned perceptual image patch similarity (LPIPS) \cite{zhang2018unreasonable} to evaluate the visual quality of rendered frames. We do not involve the quantitative scores of MoFaNeRF, considering the rendered 3D faces of MoFaNeRF significantly differ from the video frames.

\subsection{Reconstruction and Novel View Synthesis}
\label{sec:exp_rec}

We first evaluate our FDNeRF and baseline methods on the task of 3D face reconstruction and novel view synthesis based on few-shot dynamic frames. Specifically, only three dynamic frames extracted from a monocular talking video are used for 3D face modeling. Then, three novel views (frontal, left and right sides) with facial expression same as the first input frame are synthesized (if suitable). Moreover, since NeRF, NeRFace, and HyperNeRF require dense input views for 3D modeling, we add additional 27 frames uniformly extracted from the video to improve their performance and denote them as NeRF$_{30}$, NeRFace$_{30}$, and HyperNeRF$_{30}$, respectively.

As shown in Fig. \ref{fig:rec} and Table \ref{tab:metric}, compared to these baselines, our FDNeRF achieves more realistic 3D face reconstruction with view-consistent facial expressions. Specifically, NeRF with three dynamic frames as input fails to produce clear 3D faces. With more inputs, NeRF$_{30}$ enables inference of rough facial contours but cannot capture the misaligned facial details since NeRF is proposed for modeling static scenes. Compared to the vanilla NeRF, NeRFace and HyperNeRF are more sensitive to the number of input frames due to their complex designs for dynamic modeling. They cannot estimate reasonable 3D facial structures with only three dynamic frames since they require abundant inputs to generalize the facial expression space or the 3D deformation field. Even if we try to increase the input frames as in NeRFace$_{30}$ and HyperNeRF$_{30}$, such collapse still happens. PixelNeRF produces more plausible 3D faces than the previous methods. Nonetheless, as it is designed to reconstruct static scenes, PixelNeRF fails to distinguish the expression misalignment between frames, resulting in blurred textures and inaccurate facial expressions. For MoFaNeRF, we observe that it strongly overfits the property of training data in terms of occlusion, lighting, and facial shapes, which could not generalize to the video data well. By contrast, with the well-designed CFW module, our FDNeRF eliminates inconsistencies among input frames and reconstructs 3D faces with the desired expression and realistic facial details.

\subsection{Expression Editing} % exp-driven, video-driven
\label{sec:exp_edit}
Thanks to the well-designed CFW module, our FDNeRF enables editing 3D faces to novel expressions beyond those in input frames. Since most NeRF-based methods cannot accomplish the expression editing task, we only compare with NeRFace and MoFaNeRF in this section. Both approaches are conditioned on explicit 3DMM expression semantics and theoretically support expression editing of 3D faces. 
Considering that NeRFace fails to reconstruct a basic facial contour based on three dynamic frames, we adopt faces generated by NeRFace$_{30}$ on the editing task. As shown in Fig. \ref{fig:exp_edit}, we use five specific expressions as the target to drive the 3D faces.
NeRFace fails to perform expression editing since it cannot to generalize the facial expression space and perceive expression changes without enough input frames.
Although MoFaNeRF fails to fit reasonable 3D faces, it allows editing facial expressions to a certain extent. By contrast, our FDNeRF could faithfully edit the expression and render realistic results based on both general priors learned from the training stage and personal information extracted from three input frames. As a result, our expression edit, like "mouth open" in Fig. \ref{fig:exp_edit}, is not with the same deformation for all persons but adaptive to different individuals.

\subsection{Extension to Video-driven Reenactment.}~~ 
\label{sec:exp_exten}
\begin{figure}
  \includegraphics[width=1\linewidth]{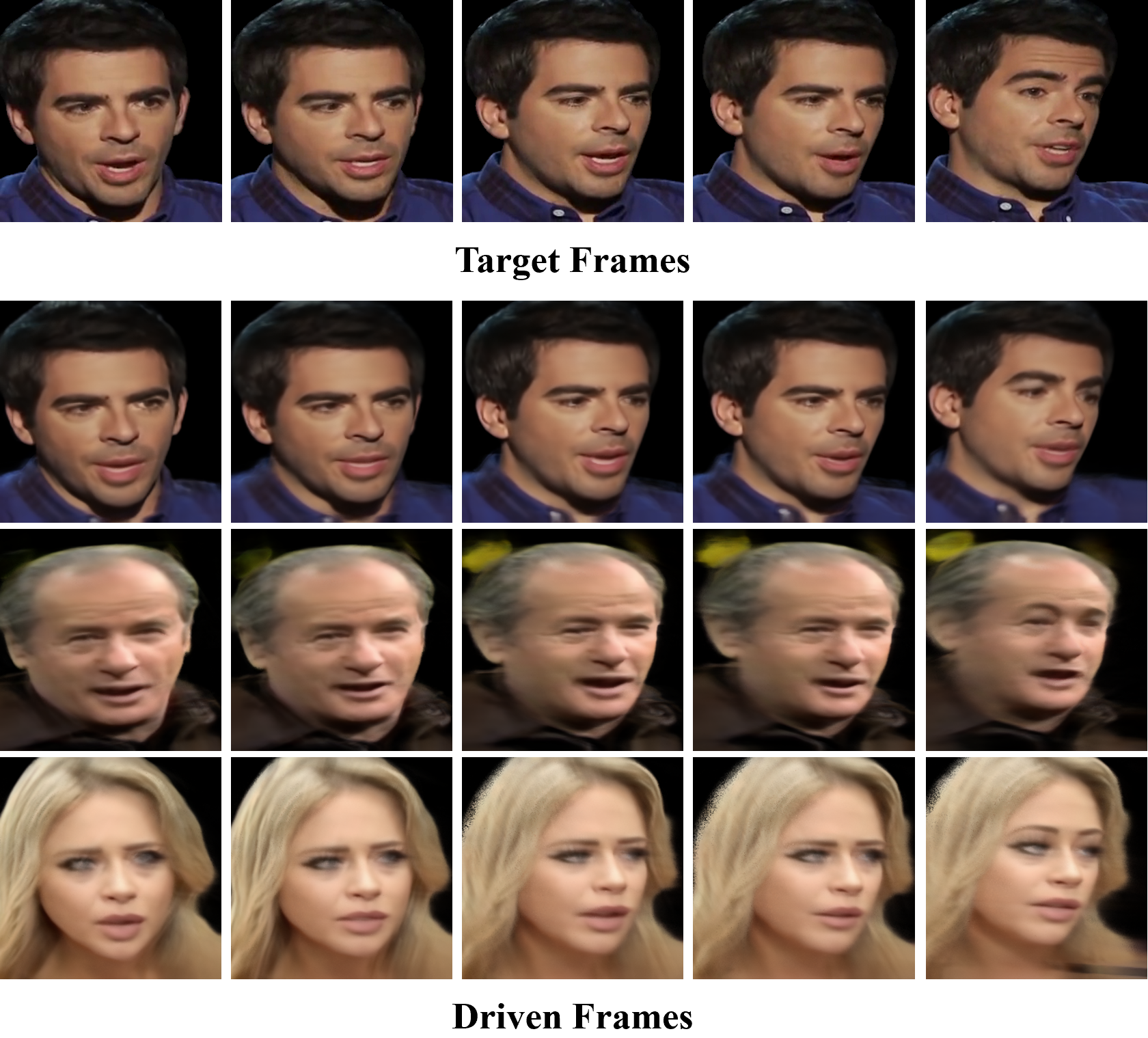}
  \vspace{-0.2in}
  \caption{Extension on video-driven reenactment. Input images are from the VoxCeleb dataset \cite{nagrani2017voxceleb}.}
  \label{fig:video}
  \vspace{-0.1in}
\end{figure}

\begin{figure}
  \includegraphics[width=1\linewidth]{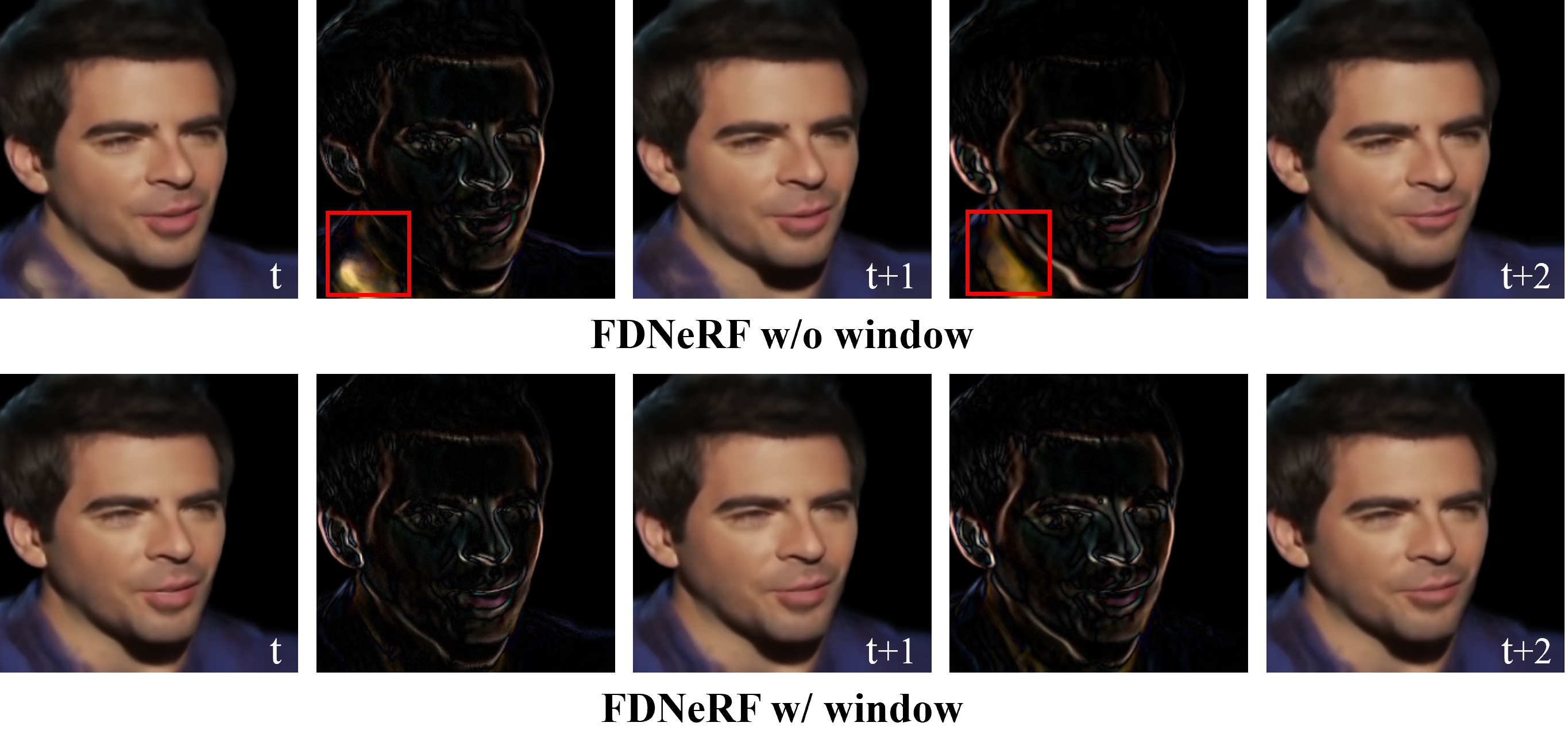}
  \vspace{-0.2in}
  \caption{Effectiveness of parameter window. Input images are from the VoxCeleb dataset. \cite{nagrani2017voxceleb}.}
  \label{fig:video_arti}
  \vspace{-0.1in}
\end{figure}

Since our FDNeRF supports expression editing, it could be applied to video-driven 3D reenactment by using the expression parameters of a video sequence. 
However, in practice, we find that slight inconsistencies between the estimated parameters of adjacent frames may cause the discontinuity artifacts in the reenactment results (shown in Fig.~\ref{fig:video_arti}), although the frame-to-frame tracking method \cite{thies2016face2face} can reduce the temporal jitter to some extent.
To alleviate this issue, we modify the semantic mapping network to receive a set of parameters instead of a per-frame one to output a latent code. With this modification, we input parameters of a window with continuous frames as the target expression semantic, where the parameter window is set to the forward and backward $L$ frames centered on the target frame. $L$ is set to 13 in our video-driven experiments. The effectiveness of the parameter window is shown in Fig.~\ref{fig:video_arti}, and Fig.~\ref{fig:video} gives more reenactment results driven by a video of the same person or a different person.

% p1-[start] ---------- move to the supplementary material -----------
% \begin{figure}
%   \includegraphics[width=1\linewidth]{figures/PNGs/abla_warp.png}
%   \vspace{-0.2in}
%   \caption{Ablation study on warping strategy. Input images are from the VoxCeleb dataset \cite{nagrani2017voxceleb}.}
%   \label{fig:warp}
% \end{figure}
% p1-[end]--------------------------------------------------------

\section{Conclusion}
In this paper, we propose the FDNeRF for 3D face reconstruction and expression editing based on a small number of dynamic frames extracted from a talking head video. We design the expression-conditioned feature warping module to eliminate inconsistencies between dynamic frames and a radiance field reconstruction module to perform accurate 3D reconstruction with aligned features. Consisting of these well-designed modules, the proposed FDNeRF demonstrates superior performance on novel view synthesis and arbitrary expression editing tasks. We further extend the FDNeRF with a window-based strategy for temporal coherent video-driven reenactment.

\noindent\textbf{Limitation.}~~ Although our FDNeRF can effectively handle the expression inconsistencies between input frames and reconstruct realistic 3D faces, there are still some limitations. For example, inconsistencies in the non-face region (e.g., hair and torso), which are not conditioned on expressions, will cause some blurriness in the result, as shown in the third example of Fig.~\ref{fig:exp_edit}. It might be alleviated by introducing separate warping fields for different parts, such as a body warping field conditioned on skeleton pose \cite{su2021nerf}. Besides, the lighting inconsistency among input frames may also cause reconstruction to fail. To overcome this issue, a reflectance field \cite{srinivasan2021nerv} may be employed to decompose the illumination and achieve relighting. We seek to solve these challenges in the following works.

\noindent\textbf{Potential Social Impact.}~~ Although not the purpose of this work, our expression editing technology could be misused in some deepfake applications like fake talking video generation. The risk can be effectively relieved by existing forgery detection methods like \cite{wang2020cnn,zhao2021learning}.

% DO NOT INCLUDE ACKNOWLEDGMENTS IN AN ANONYMOUS SUBMISSION TO SIGGRAPH 2019
% \begin{acks}
% We thank anonymous reviewers for their constructive comments. This work was supported by the
% Shenzhen Basic Research General Program under Grant
% JCYJ20190814112007258.

% \end{acks}

% Bibliography
\bibliographystyle{ACM-Reference-Format}
\bibliography{bibliography}



\section*{Supplementary material}

# FDNeRF: Few-shot Dynamic Neural Radiance Fields for Face Reconstruction and Expression Editing (*Supplementary Material*)

## 1 OVERVIEW

In this supplemental material, we provide more details and additional experimental results of our method, including:

- Detailed network architectures (Section 2);
- Implementation details about data pre-processing, training, and inference procedures (Section 3);
- Comparison with 2D reenactment methods (Section 4.1);
- Ablation study on different warping strategies (Section 4.2);
- Ablation study on the number of input frames (Section 4.3);
- Ablation study on semantic mapping network (Section 4.4);
- Ablation study on face tracking errors (Section 4.5).

## 2 NETWORK ARCHITECTURES

Our conditioned feature warping (CFW) module consists of three sub-networks: a feature encoding network, a semantic mapping network, and a conditional warping network. We omitted the network details to keep the paper compact. Here, we introduce more details about the architectures of these networks.

**Feature encoding network.** The encoding network is designed with a ResNet34 backbone, as shown in Fig. 1. We extract feature maps of the first four layers, upsample them to the same resolution using bilinear interpolation, and concatenate them to form the 512-dimension feature volume.

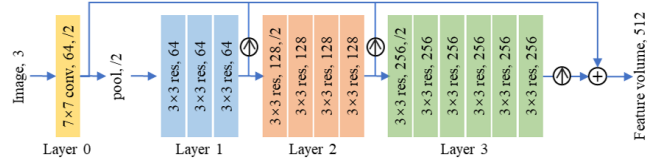

Fig. 1. Structure of the feature encoding network, where the up arrow indicates the upsampling process.

**Semantic mapping network.** As shown in Fig. 2(a), the mapping network of our original FDNeRF is structured with a 3-layer MLP. To achieve the video-driven reenactment task, we modify the mapping network to receive a set of parameters (parameter window) instead of a per-frame one to output the latent code, as shown in Fig. 2(b).

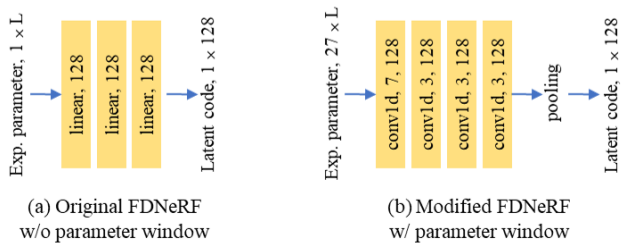

Fig. 2. Structure of the semantic mapping network. Here $L$ indicates the dimension of the expression parameter, which is 85 in all our experiments.

**Conditional warping network.** As shown in Fig. 3, the warping network is designed with an auto-encoder architecture similar to [Ren et al. 2021], which takes the feature volume and motion descriptor as input and estimates a flow field indicating the coordinate offsets between the source and target feature volumes. According to the output flow field, the warped feature volume can be calculated by warping the source volume.

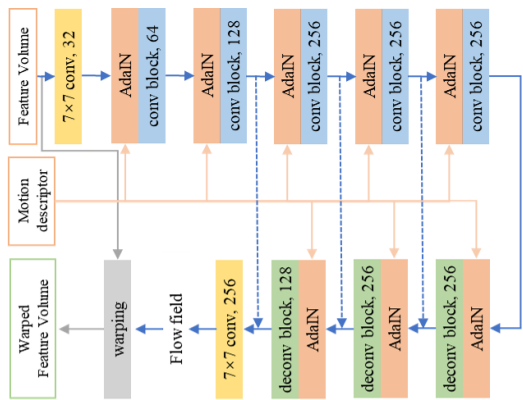

Fig. 3. Structure of the conditional warping network.

**Reconstruction network.** As shown in Fig. 4, we adopt a NeRF structure similar to PixelNeRF [Yu et al. 2021], which consists of 5 fully-connected ResNet blocks with a width of 512. To enable arbitrary input numbers of views, we extract the feature vectors $\{\mathbf{v}_1, \ldots, \mathbf{v}_M\}$ from each input view and feed them, along with the encoded position $\gamma(\mathbf{x})$, into the first three blocks, respectively. Then, we perform an average-pooling operation among all input views after the third block and obtain an aggregated intermediate vector for the remaining two blocks. Unlike PixelNeRF [Yu et al. 2021], which feeds the view direction component $\gamma(\mathbf{d})$ at the beginning of the network, we input it into the penultimate layer to eliminate the geometric discrepancy between different views.

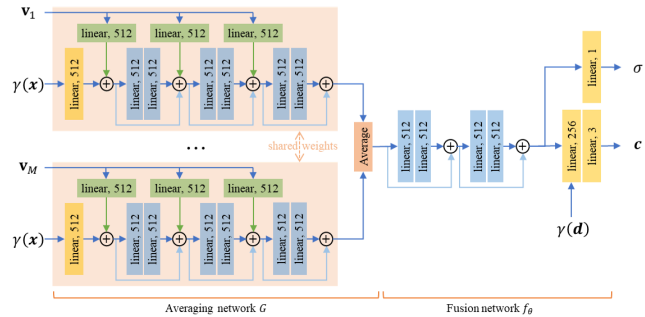

Fig. 4. Structure of the reconstruction network.

## 3 IMPLEMENTATION DETAILS

**Data pre-processing.** We randomly select 213 talking videos from the VoxCeleb dataset [Nagrani et al. 2017], and we find that this

1

amount of data is sufficient to fit our model. Then, we follow the pre-processing method described in [Siarohin et al. 2019] to crop faces within a 256×256 bounding box. Following AD-NeRF [Guo et al. 2021], we employ the face tracking method [Thies et al. 2016] to estimate the expression semantic, corresponding face pose, and intrinsic matrix for each frame of the cropped video. Here, the face tracking method can be replaced by other similar methods like Deep3DFace [Deng et al. 2019], 3DDFA [Guo et al. 2020], and pre-processing method of MoFaNeRF [Zhuang et al. 2021]. Theoretically, the more accurate the estimation of the expression parameters and face poses, the better the performance of our method.

**Training procedure.** During the training stage, we first randomly select one frame from one of the training videos as the target view, then randomly select 1-12 frames from the remaining frames of the same video as the input views. The corresponding expression parameters and poses are fed into our framework to reconstruct the 3D face model and produce the rendered view. The L2 loss between the pixel colors of rendered and target views is employed to update the networks' weights, including the encoding, mapping, warping, and reconstruction networks. For the positional encoding used on the position and direction of each query point, we follow the definition in NeRF [Mildenhall et al. 2020] and set the degrees of position and direction to 6 and 4, respectively. Note that, there are no parameters that need to be updated in the positional encoding during the training procedure.

**Inference procedure.** To quantitatively evaluate the performance of our method, we randomly select one frame as the target view from one of the test videos and randomly select 1-12 frames as the input views from the remaining video frames. Such a selection strategy means that the facial information of the modeled object cannot be guaranteed to be fully covered in the input frames, making the uncovered viewpoints not well modeled. To model a complete 3D face in practical applications, we propose that the input frames fully cover the target face, i.e., the input frames should be captured from different viewpoints.

## 4 EXPERIMENTS

In this section, we provide additional experimental results on video-driven reenactment and ablation studies.

### 4.1 Comparison with 2D reenactment methods

To further illustrate the performance of our FDNeRF in video-driven reenactment, we compare our method with two 2D reenactment methods, i.e., LSR [Meshry et al. 2021] and PIRender [Ren et al. 2021]. As shown in Fig. 6, since 2D video-driven methods only generate target frames from the 2D image domain, they lack constraints in 3D space, which makes these methods unable to ensure the view consistency of the generated results, especially for the large pose variations. In contrast, our FDNeRF is able to generate view-consistent 3D faces with desired expressions, which makes our reenactment results more realistic and visually harmonious than 2D methods.

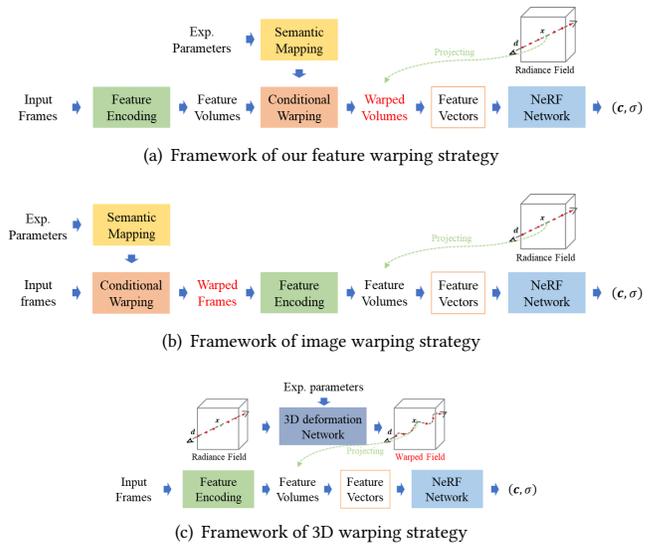

Fig. 5. Illustration of different warping strategies.

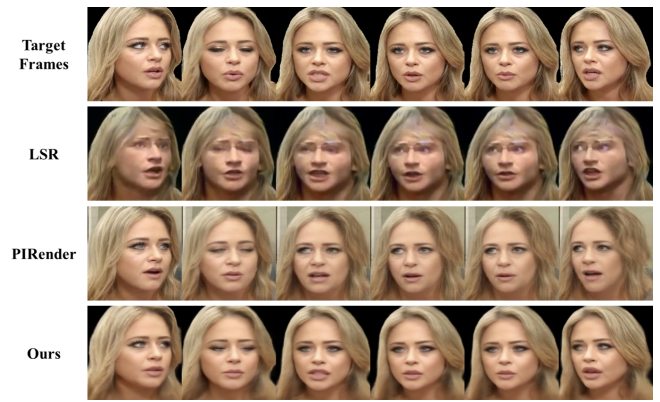

Fig. 6. Comparison with 2D reenactment methods. Images are from the VoxCeleb dataset [Nagrani et al. 2017].

### 4.2 Ablation study on different warping strategies

As described in Sec.3.2 of the paper, in addition to our feature warping strategy, there are two potential warping strategies that can be used to achieve alignment for dynamic input frames. One strategy is to employ a 3D deformation field conditioned on expression parameters like previous dynamic NeRFs [Park et al. 2021a,b]. In this 3D-warp strategy, a 3D deformation network is employed to achieve spatial warping between observation space and canonical space, as shown in Fig. 5(c). Another strategy is similar to our feature warping strategy, conducting 2D warping but at the image level, i.e., directly warping the input frames instead of feature volumes, as shown in Fig. 5(b).

To validate the effectiveness of our warping strategy used in the conditioned feature warping (CFW) module, we conduct a comparison experiment by replacing the CFW module with the two above

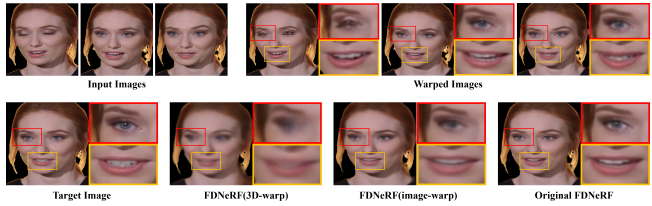

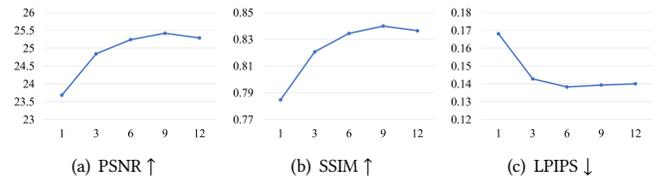

Fig. 7. Ablation study on warping strategy. Input images are from the VoxCeleb dataset [Nagrani et al. 2017].

Table 1. Quantitative results of ablation study on different warping strategies.

| Methods | PSNR ↑ | SSIM ↑ | LPIPS ↓ |
| --- | --- | --- | --- |
| FDNeRF(3D-warp) | 21.505 | 0.706 | 0.266 |
| FDNeRF(image-warp) | 24.026 | 0.797 | 0.200 |
| FDNeRF(original) | **24.847** | **0.821** | **0.142** |

strategies, denoted as FDNeRF(3D-warp) and FDNeRF(image-warp), respectively. We use the same data and optimization strategy as our original FDNeRF to train and test these two models. They are also jointly trained with the radiance field and thus under the 3D geometry constraints. As shown in Fig. 7 and Table 1, the performance of FDNeRF(3D-warp) is significantly lower than the other two 2D warping strategies since the 3D deformation field with a higher dimension is harder to learn and requires more input frames. Moreover, without seeing the whole image, the 3D deformation field defined on individual positions cannot handle identity differences. Unlike FDNeRF (3D-warp), FDNeRF (image-warp) and the FDNeRF learning 2D warping field for a whole image can converge effectively. But, FDNeRF(image-warp) warping at the image level still introduces artifacts like blurriness and inconsistency (see warped images in Fig. 7) that cause the performance degradation. In contrast, our 2D feature mapping is more robust to small warping errors and generates much sharper results.

### 4.3 Ablation study on the number of input frames

To illustrate the robustness of our method on the number of input frames, we report the variation curve of quantitative evaluation results under different input frames in Fig. 8. With the increase in the number of input frames, the performance of our method gradually improves, and becomes saturated when it reaches around nine frames because less information can be provided from additional views. Moreover, due to the random selection strategy of the input frames during training and testing, the performance may not be guaranteed to improve steadily as the views increase after saturation, as shown in Fig. 8.

### 4.4 Ablation study on semantic mapping network

To illustrate the effectiveness of our semantic mapping network, we conduct an experiment by removing the mapping network from our framework and employing the original expression parameters to guide the subsequent conditional warping network. As shown in Fig. 10 and Table 2, FDNeRF with a mapping network achieves

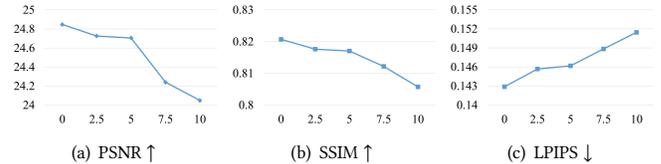

Fig. 8. Metrics under different numbers of input images.

Fig. 9. Quantitative results under different degrees of errors.

Table 2. Quantitative results of semantic mapping network.

| Methods | PSNR ↑ | SSIM ↑ | LPIPS ↓ |
| --- | --- | --- | --- |
| FDNeRF w/o mapping network | 24.743 | 0.811 | 0.180 |
| FDNeRF w/ mapping network | **24.847** | **0.821** | **0.142** |

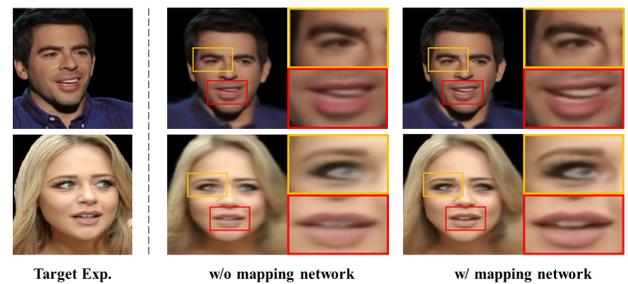

Fig. 10. Ablation study on semantic mapping network. Images are from the VoxCeleb dataset [Nagrani et al. 2017].

better performance than the one without a mapping network in both qualitative and quantitative results. Actually, the mapping network maps the original expression parameters into a higher dimensional latent space, thus allowing more representational capacity in the condition and leading to better warping results.

### 4.5 Ablation study on face tracking errors

To evaluate the robustness of our method to errors in face tracking parameters, we add $a\%$ amplitude Gaussian noise to the tracking parameters and test the performance of our method. We show the variation curve of quantitative evaluation results under different degrees of error in Fig. 9, where the value of $a$ increases from 0 to 10. Although the performance of our method degrades with the increase of noise, this performance change caused by the error of the tracking parameters is acceptable within a certain range, especially when the error is less than 5% amplitude. In other words, the performance of our method is robust to the face tracking error.



\end{document}